\documentclass[conference]{IEEEtran}
\IEEEoverridecommandlockouts
\usepackage{cite}
\usepackage{amsmath,amssymb,amsfonts}
\usepackage{algorithmic}
\usepackage{graphicx}
\usepackage{textcomp}
\usepackage{xcolor}
\usepackage{times}
\usepackage{soul}
\usepackage{url}
\usepackage[hidelinks]{hyperref}
\usepackage[utf8]{inputenc}
\usepackage[small]{caption}
\usepackage{epsfig}
\usepackage{float}
\usepackage{graphicx}
\usepackage{amsmath}
\usepackage{amsthm}
\usepackage{booktabs}
\usepackage{algorithm}
\usepackage{algorithmic}
\usepackage{multirow}
\usepackage[switch]{lineno}
\def\BibTeX{{\rm B\kern-.05em{\sc i\kern-.025em b}\kern-.08em
    T\kern-.1667em\lower.7ex\hbox{E}\kern-.125emX}}
\begin{document}

\title{MU-MAE: Multimodal Masked Autoencoders-Based One-Shot Learning
}

\author{\IEEEauthorblockN{1\textsuperscript{st} Rex Liu}
\IEEEauthorblockA{\textit{Department of Computer Science} \\
\textit{University of California, Davis}\\
Davis, USA \\
rexliu@ucdavis.edu}
\and
\IEEEauthorblockN{2\textsuperscript{nd} Xin Liu}
\IEEEauthorblockA{\textit{Department of Computer Science} \\
\textit{University of California, Davis}\\
Davis, USA \\
xinliu@ucdavis.edu}
}

\maketitle

\begin{abstract}
With the exponential growth of multimedia data, leveraging multimodal sensors presents a promising approach for improving accuracy in human activity recognition. Nevertheless, accurately identifying these activities using both video data and wearable sensor data presents challenges due to the labor-intensive data annotation, and reliance on external pretrained models or additional data. To address these challenges, we introduce Multimodal Masked Autoencoders-Based One-Shot Learning (Mu-MAE). Mu-MAE integrates a multimodal masked autoencoder with a synchronized masking strategy tailored for wearable sensors. This masking strategy compels the networks to capture more meaningful spatiotemporal features, which enables effective self-supervised pretraining without the need for external data. Furthermore, Mu-MAE leverages the representation extracted from multimodal masked autoencoders as prior information input to a cross-attention multimodal fusion layer. This fusion layer emphasizes spatiotemporal features requiring attention across different modalities while highlighting differences from other classes, aiding in the classification of various classes in metric-based one-shot learning. Comprehensive evaluations on MMAct one-shot classification show that Mu-MAE outperforms all the evaluated approaches, achieving up to an 80.17\% accuracy for five-way one-shot multimodal classification, without the use of additional data.
\end{abstract}


\section{Introduction}
Human Activity Recognition (HAR) plays a pivotal role in design and deployment of intelligent systems across various domains, ranging from healthcare and assistive technologies to smart homes and autonomous vehicles \cite{sensor_overview,bwcnn,smarthome,autocar}. For instance, precise activity recognition can facilitate collaborative robots in assisting workers by delivering tools at the right moment~\cite{robot}. 

In the last decade, extensive research in the field of HAR has been fueled by the spread of smart devices equipped with built-in wearable sensors, high-resolution visual devices, and advancements in artificial intelligence technology. This research has predominantly centered around the use of unimodal sensor data, such as wearable sensors~\cite{ICUpaper,sensor_ad,sensor_deep} and visual inputs~\cite{trx,vivit,mastaf}. However, unimodal algorithms encounter difficulties in certain real-world scenarios, especially when it comes to distinguishing similar activities using a single modality, such as differentiating between carrying a light and a heavy object~\cite{mumu}. As a response, incorporating additional modalities to support activity identification and improve overall accuracy has become a viable and increasingly popular direction.

Multimodal HAR aims to develop models capable of processing and correlating information from various modalities\cite{hamlet}. The use of multimodal representation is anticipated to enhance the performance of human activity recognition, as each modality has the potential to capture distinct and complementary information. Nevertheless, existing multimodal learning approaches face two critical challenges in real-world settings.

Firstly, the collection and annotation of multimodal data is labor-intensive as the number of data modalities increases. Specifically, when transferring a pretrained multimodal model to a target dataset with vision modalities, there is a requirement for a noteworthy amount of annotated multimodal data pertaining to the novel classes in the target dataset for fine-tuning. In the absence of sufficient labeled multimodal data, the performance of multimodal classification is likely to decline.

Secondly, since many multimodal approaches contain large-scale models like ResNet~\cite{resnet} and transformers~\cite{attention}, especially when dealing with high-dimensional data such as videos, external pretrained models or extra data are necessary for model pretraining~\cite{timesformer,vivit}. Without external pretrained models or extra data, most multimodal approaches may produce unsatisfactory results, consequently diminishing their applicability in multimodal scenarios.

\begin{figure*}[!ht]
\centering
\includegraphics[width=0.63\textwidth]{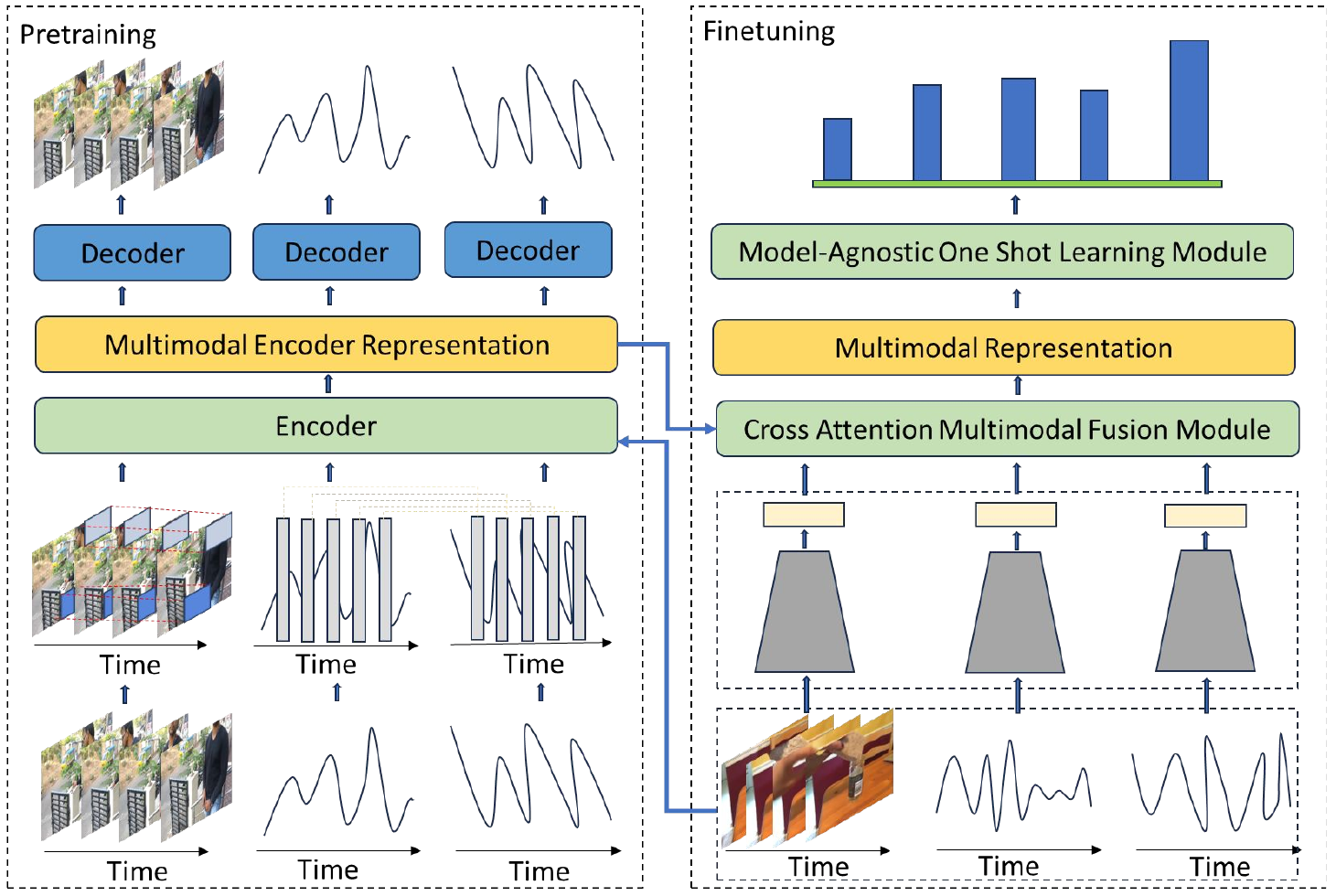}
\caption{Illustration of the Multimodal Masked Autoencoders-Based One-Shot Learning (MU-MAE), involving a video modality and two time series modalities. The MU-MAE framework involves two steps.
In the first step, known as the pretraining process, a tube masking strategy is employed, and then we get representations of unmasked video data, inspired by the VideoMAE framework~\cite{videomae}. Simultaneously, a synchronized masking strategy is applied to the other two physical sensor modalities. This synchronized masking strategy entails masking all time series data at the same specific time points. The concatenated representation, including position information, is then fed into the encoder module to produce the multimodal encoder representation. Subsequently, individual decoders are trained for each modality using mean square error loss to reconstruct the respective modality data. The second step involves a finetuning process focused on one-shot multimodal classification. Unimodal feature encoders pretrained in the pretraining process are applied to extract unimodal representations. The unimodal representations and the multimodal encoder representations are fed into the cross attention multimodal fusion module.  This process produces the multimodal representation, which is then directed into the model-agnostic one-shot learning module for classification.
More details can be found in Section~\ref{s:method}.
}
\label{workframe}
\end{figure*}

To overcome these challenges, we present a novel approach named Multimodal Masked Autoencoders-Based One-Shot Learning (Mu-MAE)(as shown in Fig~\ref{workframe}). Firstly, we introduce one-shot multimodal learning to significantly reduce the annotation cost associated with multimodality data. In one-shot multimodal classification, the multimodal samples in the training and test sets come from different classes, specifically unseen classes in the test set. The objective of a one-shot multimodal classification model is to classify an unlabeled multimodality sample (query) to the unseen class (support set). Secondly, for efficient multimodal model pretraining without relying on external data or pretrained models, we propose a multimodal masked autoencoders with a synchronized masking strategy for wearable sensor data. These masking strategies compel the networks to capture more meaningful spatiotemporal features, thereby making multimodal masked autoencoders a more intricate yet rewarding self-supervised learning task. Lastly, we utilize the multimodal representation extracted from multimodal masked autoencoders as prior information input to the cross-attention multimodal fusion layer. This fusion layer highlights spatiotemporal features that require attention across different modalities while emphasizing differences from other classes. Further details can be found in Section~\ref{s:method}.

\noindent 
\textbf{Contributions.} We make the following contributions. 

\noindent
1.We present Mu-MAE, an effective and efficient one-shot classification model guided by multimodal masked autoencoders. Mu-MAE possesses the capability to train a vanilla multimodal model directly on the multimodality dataset without relying on any pretrained model or external multimodal data.

\noindent
2.
We design a fusion mechanism that integrates cross-attention networks with the input of multimodal representation learned from the task of reconstructing multimodal data. This integration significantly augments the crucial spatial and temporal regions within the multimodal representation, contributing to the efficacy of the one-shot learning architecture.

\noindent 
3.
We conduct a thorough evaluation of Mu-MAE on the one-shot data split of MMAct~\cite{mmact}, alongside recent multimodal approaches, namely HAMLET~\cite{hamlet} and MuMu~\cite{mumu}. In comparison to these existing works, Mu-MAE enhances state-of-the-art performance without the need for any pretrained model or additional data, achieving 80.17\% for five-way one-shot classification.
Our code and the one-shot data split of MMAct are available at \url{https://anonymous.4open.science/r/mu-mae-CAC4}.

\section{Related work}
\noindent
\textbf{Multimodal classification.}
Earlier multimodal learning approaches primarily focused on extracting representations from similar modalities~\cite{early_multimodal_work1,early_multimodal_work2,early_multimodal_work3}. For instance, the two-stream CNN excelled at capturing spatial and temporal features from visual data~\cite{early_multimodal_work2}, while Feichtenhofer's two-stream learning model varied data sampling rates to extract spatial-temporal features~\cite{slowfast}.
Recent research underscores the development of multimodal learning methods that effectively leverage complementary features from different modalities to overcome dependencies on single-modality data in modality-specific HAR models. For example, in ~\cite{long_work}, they first use attention model to extract unimodal features, which are then fused to generate multimodal representations. Challenges persist in efficiently fusing various unimodal features, leading to the exploration of different fusion approaches, including early fusion, late fusion, and hybrid fusion strategies~\cite{early_multimodal_work2,slowfast,late_fusion}. Simonyan et al.'s two-stream CNN architecture~\cite{early_multimodal_work2}, incorporating spatial and temporal networks, has been extensively studied and proven effective in recent works, employing residual connections~\cite{residule_conncet} and slow-fast network techniques~\cite{slowfast}. Other investigations focus on simultaneous feature fusion from diverse modalities, such as video, and wearable sensor modalities. 
HAMLET~\cite{hamlet} employs a hierarchical architecture with a multi-head self-attention mechanism to encode spatio-temporal features from unimodal data in the lower layer and then fuse them in upper layer. 
MuMu~\cite{mumu} incorporates an auxiliary task involving activity group classification to guide the fusion with unimodal representations.
Despite these advancements, the ongoing challenge in the field lies in dynamically selecting unimodal features to generate multimodal features.
In our Mu-MAE, we address this challenge using a cross-attention multimodal fusion module, dynamically highlighting spatiotemporal features that require attention across different modalities while emphasizing differences from other classes.

\noindent
\textbf{Masked visual modeling,}
Masked visual modeling has proven to be a robust strategy for acquiring impactful representations by employing a sequential process involving masking and subsequent reconstruction. Although early efforts predominantly concentrated on the image domain, employing techniques such as denoised autoencoders~\cite{ecr} and convolutions for inpainting missing regions~\cite{context}, recent advancements have expanded the scope of this methodology to encompass videos. Vision transformer architectures like BEiT~\cite{beit}, BEVT~\cite{bevt}, and VIMPAC~\cite{vimpac} were inspired by language models~\cite{language_few,bert}, opting for the prediction of discrete tokens to glean visual representations from both images and videos. The introduction of MAE~\cite{mae} brought forth an asymmetric encoder-decoder architecture finely tuned for masked image modeling, whereas VideoMAE~\cite{videomae} took a distinctive approach by directly reconstructing pixels in a more straightforward yet highly effective video masked autoencoder. The evolution in masked visual modeling signals a notable shift towards direct pixel-level reconstruction, enhancing self-supervised pretraining in both image and video domains. 
Subsequently,  VideoMAE V2~\cite{videomae_v2} introduced an effective pretraining method utilizing a dual masking strategy. In this approach, an encoder processes a subset of video tokens, and a decoder manages another subset of video tokens. This strategy facilitates the efficient pretraining of billion-level models in the video domain.

\noindent
\textbf{One-shot learning,}
One-shot learning algorithms are typically classified into three primary categories: optimization-based methods~\cite{MAML}, model-based methods~\cite{Meta-Networks,memory-modelbased}, and metric-learning-based methods~\cite{trx,prototypical,ITANet,mastaf}. Among these, metric-learning-based methods emerge as particularly promising, as evidenced by their superior performance in prior studies~\cite{trx,ITANet,mastaf}.
Metric-learning-based approaches compute the distance between representations of support and query samples, utilizing the nearest neighbor for classification. The fundamental principle involves maintaining closeness between representations of similar classes while ensuring differentiation between representations of dissimilar classes. For instance, MASTAF~\cite{mastaf} highlights spatio-temporal features that demand attention for each class, simultaneously accentuating distinctions from other classes. In our Mu-MAE, we also leverage the metric-learning-based method. To optimize its effectiveness, we employ a cross-attention multimodal fusion module to enhance the differentiation of each class's spatio-temporal features, contributing to improved performance in one-shot classification tasks.

\section{Proposed Method}
\label{s:method}
\subsection{Problem definition}
A $C$-way one-shot multimodal learning problem involves learning multimodal representation for model-agnostic one-shot learning, where $C$ denotes the number of categories in the support set. Similar to the one-shot learning problem, we aim to recognize a set of multimodal data into one of given annotated categories, by assessing the similarity between pairs of multimodal representations ($R^{m}$) from $N$ heterogeneous modalities, where $N$ denotes the number of modalities.


We use $X^{r}=\{X^{r}_{1}, ...,X^{r}_{i},...,X^{r}_{N}\}$ to denote the raw feature of $N$ heterogeneous modalities and $X^{r}_{i}$ stands for raw feature of $i$ modality.
The final goal is to get $R^{m}$ from $X^{r}$ and then predict $R^{m}$ to one of the classes.
\subsection{Approach Overview}
Our proposed Multimodal Masked Autoencoders-Based One-Shot Learning consists of four learning modules (as shown in Fig~\ref{workframe}):

\noindent 
\textbf{Unimodal Embedding module} extracts the representation for each modality.

\noindent 
\textbf{Multimodal Masked Autoencoders} engages in the pretraining of each Unimodal representation encoder, extracting the multimodal representation from both video and physical sensor modalities, which serves as the q value in the fusion layer during the finetuning process.

\noindent 
\textbf{Cross attention multimodal fusion module} integrates representations of all modalities through the utilization of cross-attention mechanisms.

\noindent 
\textbf{Model-agnostic one-shot learning module} classifies a query multimodal instance based on the similarity between the representation of the query and the representation of each class in the support set.

\subsection{Unimodal Embedding module} 
In the Mu-MAE model, the goal of the unimodal embedding module is to learn the spatio-temporal representations for each modality. We use $f^\varphi_{m}=\{f^\varphi_{1}, ...,f^\varphi_{i},...,f^\varphi_{N}\}$ to denote the unimodal embedding module of $N$ heterogeneous modalities and $f^\varphi_{i}$ stands for unimodal embedding module of $i$ modality.

Given a raw feature extracted from the $i$ modality, $X^{r}_{i}$, let $U_i$ denote the representation learned from the unimodal embedding module:
\begin{equation}
U_i = f^\varphi_{i}(X^{r}_{i}). \label{feature_extract}
\end{equation}
\subsection{Multimodal Masked Autoencoders}
The multimodal masked autoencoders module has two primary objectives. Firstly, it aims to train a vanilla unimodal embedding network for each modality directly on the multimodal dataset, without any pretrained models. 
Secondly, the multimodal representations extracted from the reconstruction task serve as prior information for the efficient fusion of multimodal representations in one-shot classification tasks. 

As shown in Fig~\ref{workframe}, we first adopt a tube masking strategy and then get the representation of unmasked video data $R^v_{unmask}$, inspired by the approach employed in the VideoMAE framework~\cite{videomae}. Then for other physical sensor modalities, we utilize the synchronized masking strategy to get the representations for each unmasked sensor data. We use $R^s_{unmask}=[R^{s,1}_{unmask}; ...'R^{s,i}_{unmask};...,R^{s,N-1}_{unmask}]$ to denote the concatenated representation of all the unmasked sensor data, which $R^{s,i}_{unmask}$ denotes the representation of $i$ unmasked sensor data. 
The synchronized masking strategy involves simultaneously masking all time series data at the same specific time points. This approach is helpful for mitigating information leakage during masked modeling and make masked time series data reconstruction a meaningful self-supervised pretraining task.
Then we get the concatenated representation with position information of all the unmasked modality data as:
\begin{equation}
R^m_{unmask} = [R^v_{unmask}+P^v;R^s_{unmask}+P^s], \label{multimodel_input_for_encoder}
\end{equation}
which $P^v$ and $P^s$ denote the position information of video and other sensor modalities.

\noindent 
\textbf{Multimodal Masked Autoencoders Encoder} is based on the ViT architecture~\cite{vit}, exclusively applied to visible, unmasked patches, inspired by the approach employed in the VideoMAE framework~\cite{videomae}. We use $f_{encoder}$ to denote the multimodal masked autoencoder encoder. Then we compute the representation $R^m_{encoder}$ extracted from encoder as:
\begin{equation}
R^m_{encoder} = f_{encoder}(R^m_{unmask}). \label{representation_encoder}
\end{equation}
Utilizing a masking strategy provides the benefit of training large encoders while requiring less computational resources and memory.

\noindent 
\textbf{Multimodal Masked Autoencoders Decoder} is designed in a light weight setting following the design in the video domain~\cite{videomae}, which could significantly reduces pretraining time. We use $f_{decoder}$ to denote the multimodal masked autoencoder decoder. Then we compute the representation $R^m_{decoder}$ extracted from encoder as:
\begin{equation}
R^m_{decoder} = f_{decoder}(R^m_{encoder}). \label{representation_decoder}
\end{equation}
After that, we employ the Mean Squared Error (MSE) loss as the loss function for tasks related to the reconstruction of multimodal data. 



\begin{figure}
\centering
\includegraphics[width=0.48\textwidth]{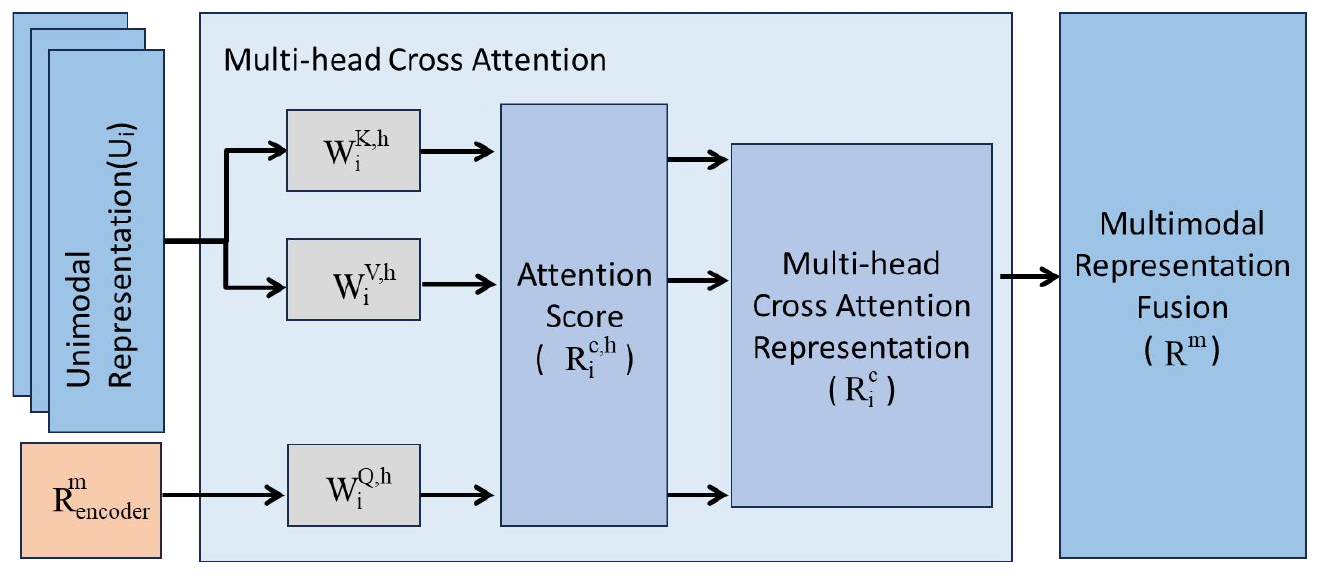}
\caption{Cross Attention Multimodal Fusion Module. $R^m_{encoder}$ is the multimodal representation from multimodal masked autoencoders' encoder. }
\label{cross-attention}
\end{figure}

\subsection{Cross attention multimodal fusion module}
We use the representations from multimodal masked autoencoders encoder as prior information, $R^m_{encoder}$, to extract multimodal representations. One benefit is to highlight spatio-temporal features that need attention across different modalities while increasing the differences from other classes. To compute the attended multimodal representation, we utilize multi-head cross attention method (as shown in Fig~\ref{cross-attention}). First, we transforms the extracted unimodal features of the $i$ modality, $U_i$ ,to generate unimodal key ($K^{u,h}_i$) and value ($V^{u,h}_i$) feature vectors for head $h$ using the following procedure:
\begin{equation}
K^{u,h}_i = U_iW^{K,h}_i; V^{u,h}_i = U_iW^{V,h}_i, \label{compute_k_v}
\end{equation}
which $W^{K,h}_i$ and $W^{V,h}_i$ are learnable parameters. Then, we transforms the extracted representations from multimodal masked autoencoders' encoder, $R^m_{encoder}$, to generate the query feature vectors $Q^{u,h}_i$ as:
\begin{equation}
Q^{u,h}_i = R^m_{encoder}W^{Q,h}_i, \label{compute_q}
\end{equation}
which $W^{Q,h}_i$ is a learnable parameter. Then we use query feature vectors $Q^{u,h}_i$ of the $i$ modality for head $h$ to generate the multimodal representation of the $i$ modality for head $h$as:
\begin{equation}
R^{c,h}_i = softmax(\frac{Q^{u,h}_i(K^{u,h}_i)^T}{exp(d^{K}_i)}V^{u,h}_i),
\end{equation}
which $d^{K}_i$ is dimension of $K^{u,h}_i$. After that, all the head multimodal representations of the $i$ modality $R^{c,h}_i$ are concatenated and projected to produce multi-head cross attention representation of the $i$ modality ($R^{c}_i$).
\begin{equation}
R^{c}_i = [R^{c,1}_i:...:R^{c,h}_i]W^{c}_i,
\end{equation}
where $W^{c}_i$ is the projection parameter.
Following that, we concatenate the multi-head cross attention representations from all modalities and employ a linear projection to generate the multimodal representation $R^m$.
\begin{equation}
R^m = W^m[R^c_1:...R^c_i:...:R^c_N],
\end{equation}
which $W^m$ is a learnable projection parameter. 
\subsection{Model-agnostic one-shot learning module}
Once we obtain $R^m$, including spatio-temporal features with knowledge across different modalities, we apply the existing model-agnostic one-shot learning module for one-shot learning task. The model-agnostic one-shot learning module assesses the distance between the representations of support samples and query sample, and classifies them with the aid of the nearest neighbor to keep similar classes close and dissimilar classes far away. We represent the $R^m$ of support class $k$ as $R^m_{s_k}$ and the $R^m$ of a query sample as $R^m_q$. The model-agnostic one-shot learning module is denoted by $f_{one-shot}$. Then we compute the one-shot learning representation of support class $k$ ($S^m_{s_k}$) and query sample ($S^m_q$) as:
\begin{equation}
S^m_{s_k} = f_{one-shot}(R^m_{s_k}),
\end{equation}
\begin{equation}
S^m_q = f_{one-shot}(R^m_q).
\end{equation}

After that, we compute the probability of predicting $S^m_q$ as the class $k$ using one-shot learning representation:
\begin{equation}
P(y=k|S^m_q) = \frac{exp(-D_{cos}(S^m_q,S^m_{s_k}))}{\sum_{j=1}^{C}exp(-D_{cos}(S^m_q,S^m_{s_j}))},
\label{distance-self-attention}
\end{equation}
where $D_{cos}$ denotes the cosine distance and $P(y=k|S^m_q)$ denotes the probability of predicting $S^m_q$ as the class $k \in \{1, 2,...,C\}$ using one-shot learning representations. $C$ represents the number of categories in the support set.

Finally, we use a negative log-probability as the loss function of the nearest neighbor classifier based on the one-shot class label:
\begin{equation}
L = -\sum_{k=1}^{C}log P(y=k|S^m_q).
\label{L-LOSS}
\end{equation}

\begin{table*}[]
\begin{center}
\begin{tabular}{lcccc}
\hline
Method                  & Backbone & Extra data       & Fusion type & Accuracy(SD)              \\ \hline
\multirow{6}{*}{HAMLET~\cite{hamlet}} & ResNet50 & no external data & sum         & 41.77(+/-0.97)            \\
                        & ResNet50 & no external data & concat      & 42.18(+/-0.97)            \\
                        & ResNet50 & ImageNet-1k      & sum         & 75.87\%(+/-0.81)          \\
                        & ResNet50 & ImageNet-1k      & concat      & 76.05\%(+/-0.83)          \\ 
                        & ViT & no external data & concat      & 40.02(+/-0.92)            \\
                        & ViT & Kinetics-400      & concat         & 78.34\%(+/-0.79)                 \\ \hline
\multirow{4}{*}{MuMu~\cite{mumu}}   & ResNet50 & no external data & concat      & 46.02\%(+/-0.98)          \\
                        & ResNet50 & ImageNet-1k      & concat      & 78.65\%(+/-0.80)          \\ 
                        & ViT & no external data & concat      & 45.16\%(+/-0.94)          \\
                        & ViT & Kinetics-400      & concat      & 80.22\%(+/-0.84)          \\ \hline
Mu-MAE                    & ViT    & no external data & concat      & \textbf{80.17\%(+/-0.78)} \\ 
Mu-MAE                    & ViT    & Kinetics-400 & concat      & \textbf{83.82\%(+/-0.77)} \\ \hline
\end{tabular}
\caption{Comparison on 5-way 1-shot benchmarks of MMAct. SD stands for standard deviation. The best performances with or without extra data are highlighted.}
\label{start-of-art-compare}
\end{center}
\end{table*}

\section{Evaluation}
\subsection{Experimental Setup}
\noindent 
\textbf{Datasets.}
There are no established one-shot data splits available for one-shot multimodal classification involving both video and wearable sensors. Thus, we undertake the random division of classes into meta-training/validation sets and a meta-testing set within the MMAct~\cite{mmact} for the few-shot multimodal classification evaluation. This data split is presented at (\url{https://anonymous.4open.science/r/mu-mae-CAC4}) for future research.

After eliminating classes lacking data from all five modalities (video, accelerometer from phone, accelerometer from watch, gyroscope, and orientation), the MMAct dataset comprises 33 activities~\cite{mmact}, with an average of over 1,000 data samples for each activity across all five modalities. These 33 activities are then split into non-overlapping sets, with 23 assigned for use as the meta-training/validation set and 10 designated for the meta-testing set.

\noindent
\textbf{Experimental Configuration.}
Following the evaluation process in state-of-the-art one-shot learning algorithms~\cite{trx,tsn,mastaf}, we evaluate the 5-way 1-shot multimodal classification task and report the average accuracy over 10,000 randomly selected episodes from the test set. We compare our results with two state-of-the-art multimodal classification algorithms, i.e., HAMLET~\cite{hamlet}, MuMu~\cite{mumu}. In particular, MuMu necessitates the categorization of all training classes into three groups, i.e., complex, simple, desk~\cite{mumu}. To accommodate this requirement, we group the 23 activities into these three distinct categories. We use ViT-B~\cite{vit} as our video embedding network, and 1-D CNN as the unimodal embedding network for data from other sensor modalities. 
We use MASTAF~\cite{mastaf} as our model-agnostic one-shot learning module, as it has demonstrated state-of-the-art performance with the model-agnostic embedding. The unimodal features from physical sensor modalities are encoded into 64-sized feature embeddings. 
During the pretraining phase of the multimodal masked autoencoders module, we extract 16 frames from each video. In the subsequent finetuning process of MASTAF~\cite{mastaf}, 8 frames are utilized. The pretraining of the multimodal masked autoencoders is conducted on 8 NVIDIA RTX A5000 GPUs, spanning 800 epochs, while the finetuning experiments for one-shot learning involve 256,000 episodes. PyTorch and DeepSpeed~\cite{deepspeed} frameworks are utilized for expedited pretraining, and finetuning is carried out using Stochastic Gradient Descent.
\subsection{Comparison with State-of-the-art Algorithms}
\label{s:c_sota}
Table~\ref{start-of-art-compare} presents a comprehensive comparison of the overall 5-way 1-shot performance against existing methods on the MMAct one-shot data split. MASTAF serves as the chosen one-shot learning module across all algorithms, and the reported average accuracy is based on 10,000 randomly selected episodes from the test set.
For the HAMLET method, we explore two fusion merge types: concatenation-based fusion and summation-based fusion. Both fusion methods are re-implemented for comparative analysis. Additionally, pretrained and trained-from-scratch ResNet 50~\cite{resnet} and ViT~\cite{vivit} are included as embedding networks for both HAMLET and MuMu. In our method, Mu-MAE, we conduct the vanilla ViT~\cite{vivit} and pretained ViT~\cite{videomae} on the Kinetics-400 as the video embeding network with varying mask ratios and decoder depths in the pretraining process, reporting the best performance achieved with an 85\% mask ratio and 4 blocks of the decoder.
As shown in Table~\ref{start-of-art-compare}, Mu-MAE without any external data outperforms trained-from-scratch state-of-the-art methods, namely HAMLET~\cite{hamlet} and MuMu~\cite{mumu},  demonstrating improvements of 40.15\% and 35.01\% in average accuracy, respectively. The lower performance of trained-from-scratch HAMLET~\cite{hamlet} and MuMu~\cite{mumu} is attributed to their inability to leverage ViT~\cite{vivit} model scale without any external data. In contrast, Mu-MAE, equipped with a multimodal masked autoencoder, effortlessly scales up with potent backbones (e.g., ViT~\cite{vivit}), attaining an accuracy of 80.17\% on MMAct~\cite{mmact} without relying on external data.

Additionally, in the one-shot multimodal learning task, Mu-MAE with externald data outperforms the pretrained HAMLET and MuMu with an imporvement of 5.48\% and 3.6\%, reprectively. 
These improvement highlight Mu-MAE's capacity to generate spatiotemporal features that demand attention across diverse modalities, simultaneously amplifying the distinctions from other classes in one-shot learning.

While Mu-MAE with pretrained ViT~\cite{videomae} exhibits superior performance compared to Mu-MAE with vanilla ViT, the slight gap in performance between these two models highlights the efficacy of our multimodal masked autoencoders in achieving good performance using only the target dataset without relying on external data. Our approach reduces computation costs and resource requirements.

\subsection{Ablation study}
As demonstrated in Section~\ref{s:c_sota}, our Mu-MAE shows better performance compared to other state-of-the-art multimodal classification algorithms in a one-shot scenario, even without any external data. We conduct in-depth ablation studies on MMAct to show the impact of each module. 

\begin{table}[]
\begin{center}
\begin{tabular}{lc}
\hline
Method               & Accuracy                             \\ \hline
Mu-MAE-scratch       & 41.21\%(+/-0.91)                     \\
Mu-MAE-without-cross & \multicolumn{1}{l}{78.46\%(+/-0.81)} \\
Mu-MAE               & \textbf{80.17\%(+/-0.78)}            \\ \hline
\end{tabular}
\caption{Comparison results with two variants Mu-MAE for 5-way 1-shot classification on MMAct. The best performance is highlighted.}
\label{mae-pre-training}
\end{center}
\end{table}


\begin{table}[]
\begin{center}
\begin{tabular}{lcc}
\hline
Mask type & Mask ratio & Accuracy                  \\ \hline
random    & 85\%       & 79.01\%(+/-0.82)          \\
synchronized    & 75\%       & 79.12\%(+/-0.77)
\\
synchronized  & 85\%       & \textbf{80.17\%(+/-0.78)} 
\\
synchronized    & 95\%       & 78.25\%(+/-0.81)
\\ \hline
\end{tabular}
\caption{Mask design. Comparison results with different mask types and mask ratios for 5-way 1-shot classification on MMAct. The best performance is highlighted.}
\label{mask_type}
\end{center}
\end{table}

\noindent
\textbf{Multimodal masked autoencoders and Cross attention multimodal fusion.}
To explore the effectiveness of the multimodal masked autoencoders, we introduce two comparative models, namely Mu-Mae-scratch and Mu-Mae-without-cross. In Mu-Mae-scratch, video representations obtained from the vanilla ViT~\cite{vit} and other physical sensor representations are concatenated and processed through a linear projection to generate the multimodal representation. Subsequently, this multimodal representation is input into the MASTAF~\cite{mastaf} one-shot learning architecture. 
Mu-Mae-without-cross is essentially the same as Mu-Mae-scratch, except for the fact that the unimodal embedding networks undergo pretraining using multimodal masked autoencoders.The comparative results are presented in Table~\ref{mae-pre-training}.

In contrast to Mu-MAE-without-cross, the inclusion of the cross-attention multimodal fusion mechanism in Mu-MAE leads to 1.71\% performance improvement. This suggests that representations obtained from multimodal masked autoencoders contain valuable spatiotemporal knowledge across various modalities. Consequently, this enriched information facilitates the fusion layer in generating more distinctive features, thereby improving differentiation from other classes. When compared to Mu-MAE-scratch, after adding the multimodal masked autoencoder pretraining process, the other two models achieve significant performance enhancements. It demonstrates that the pretraining of multimodal masked autoencoders is essential for realizing the advantages of the model scale, especially in larger-scale models such as ViT~\cite{vivit} and ResNet 50~\cite{resnet}.


\noindent
\textbf{Masking strategy and mask ratio.}
In our Mu-MAE design, we implement a synchronized masking strategy for all time series sensor data. One advantage of employing a synchronized masking strategy is to prevent information leakage between these time series sensor data. To evaluate the effect of the synchronized masking strategy, we conduct an experiment with plain random masking, setting the mask ratio at 85\%, as outlined in Table~\ref{mask_type}. The results reveal that Mu-MAE with the synchronized masking strategy outperforms its counterpart with plain random masking, achieving a 1.16\% improvement in accuracy. When increasing the masking ratio from 75\% to 85\% for synchronized masking, the performance on 5-way 1-shot multimodal classification boosts from 79.12\% to 80.17\%. However, when the masking ratio is further increased from 85\% to 95\% for synchronized masking, the performance on 5-way 1-shot multimodal classification decreases from 80.17\% to 78.25\%, which means a 85\% mask ratio is a good tradeoff for 5-way 1-shot multimodal classification on MMAct~\cite{mmact}. These outcomes demonstrate that our synchronized masking designs make the networks to capture more useful spatiotemporal features, making Mu-MAE a more challenging yet rewarding self-supervised learning task.

\section{Conclusion}
This paper proposes a Multimodal Masked Autoencoders-Based One-Shot Learning (Mu-MAE). Mu-MAE is a simple and efficient one-shot multimodal classification framework without using any extra data for pretraining. Mu-MAE makes the most of the knowledge learned from multimodal masked autoencoders and uses a cross-attention multimodal fusion module to highlight the spatiotemporal features for one-shot multimodal classification. 
Mu-MAE outperforms existing methods (HAMLET and MuMu) by achieving 80.17\% for five-way one-shot multimodal classification, without relying on pretrained models or additional data.


\bibliographystyle{unsrt}
\bibliography{mumae_ref}

\begin{thebibliography}{10}

\bibitem{sensor_overview}
Rex Liu, Albara~Ah Ramli, Huanle Zhang, Erik Henricson, and Xin Liu.
\newblock {\em An Overview of Human Activity Recognition Using Wearable Sensors: Healthcare and Artificial Intelligence}, page 1–14.
\newblock Springer International Publishing, 2022.

\bibitem{bwcnn}
Albara~Ah Ramli, Rex Liu, Rahul Krishnamoorthy, I.~B. Vishal, Xiaoxiao Wang, Ilias Tagkopoulos, and Xin Liu.
\newblock Bwcnn: Blink to word, a real-time convolutional neural network approach.
\newblock In {\em International Conference on Internet of Things (ICIOT)}, pages 133--140, 2020.

\bibitem{smarthome}
Batyrzhan~K. Akhmetzhanov, Omar~Aslan Gazizuly, Zhanserik Nurlan, and Nurkhat Zhakiyev.
\newblock Integration of a video surveillance system into a smart home using the home assistant platform.
\newblock In {\em 2022 International Conference on Smart Information Systems and Technologies (SIST)}, pages 1--5, 2022.

\bibitem{autocar}
Ekim Yurtsever, Jacob Lambert, Alexander Carballo, and Kazuya Takeda.
\newblock A survey of autonomous driving: Common practices and emerging technologies.
\newblock {\em IEEE Access}, 8:58443–58469, 2020.

\bibitem{robot}
Tariq Iqbal, Samantha Rack, and Laurel~D. Riek.
\newblock Movement coordination in human–robot teams: A dynamical systems approach.
\newblock {\em IEEE Transactions on Robotics}, 32(4):909--919, 2016.

\bibitem{ICUpaper}
Rex Liu, Sarina~A Fazio, Huanle Zhang, Albara~Ah Ramli, Xin Liu, and Jason~Yeates Adams.
\newblock Early mobility recognition for intensive care unit patients using accelerometers.
\newblock In {\em KDD Workshop on Artificial Intelligence of Things (AIoT)}, pages 1--6, 2021.

\bibitem{sensor_ad}
Ramachandran Varatharajan and Gunasekaran Manogaran.
\newblock Wearable sensor devices for early detection of alzheimer disease using dynamic time warping algorithm.
\newblock {\em Cluster Computing}, 21, 03 2018.

\bibitem{sensor_deep}
Ozlem Durmaz~Incel and Sevda~Özge Bursa.
\newblock On-device deep learning for mobile and wearable sensing applications: A review.
\newblock {\em IEEE Sensors Journal}, 23(6):5501--5512, 2023.

\bibitem{trx}
Toby Perrett, Alessandro Masullo, Tilo Burghardt, Majid Mirmehdi, and Dima Damen.
\newblock Temporal-relational crosstransformers for few-shot action recognition.
\newblock 2021.

\bibitem{vivit}
Anurag Arnab, Mostafa Dehghani, Georg Heigold, Chen Sun, Mario Lučić, and Cordelia Schmid.
\newblock Vivit: A video vision transformer.
\newblock 2021.

\bibitem{mastaf}
Xin Liu, Huanle Zhang, Hamed Pirsiavash, and Xin Liu.
\newblock Mastaf: A model-agnostic spatio-temporal attention fusion network for few-shot video classification.
\newblock 2023.

\bibitem{mumu}
Md~Mofijul Islam and Tariq Iqbal.
\newblock Mumu: Cooperative multitask learning-based guided multimodal fusion.
\newblock In {\em AAAI}, 02 2022.

\bibitem{hamlet}
Md.~Mofijul Islam and Tariq Iqbal.
\newblock Hamlet: A hierarchical multimodal attention-based human activity recognition algorithm.
\newblock {\em 2020 IEEE/RSJ International Conference on Intelligent Robots and Systems (IROS)}, pages 10285--10292, 2020.

\bibitem{resnet}
Kaiming He, Xiangyu Zhang, Shaoqing Ren, and Jian Sun.
\newblock Deep residual learning for image recognition.
\newblock In {\em CVPR}, pages 770--778, 2016.

\bibitem{attention}
Ashish Vaswani, Noam Shazeer, Niki Parmar, Jakob Uszkoreit, Llion Jones, Aidan~N Gomez, \L~ukasz Kaiser, and Illia Polosukhin.
\newblock Attention is all you need.
\newblock volume~30, 2017.

\bibitem{timesformer}
Gedas Bertasius, Heng Wang, and Lorenzo Torresani.
\newblock Is space-time attention all you need for video understanding?
\newblock 2021.

\bibitem{videomae}
Zhan Tong, Yibing Song, Jue Wang, and Limin Wang.
\newblock Video{MAE}: Masked autoencoders are data-efficient learners for self-supervised video pre-training.
\newblock In {\em Advances in Neural Information Processing Systems}, 2022.

\bibitem{mmact}
Quan Kong, Ziming Wu, Ziwei Deng, Martin Klinkigt, Bin Tong, and Tomokazu Murakami.
\newblock Mmact: A large-scale dataset for cross modal human action understanding.
\newblock In {\em The IEEE International Conference on Computer Vision (ICCV)}, October 2019.

\bibitem{early_multimodal_work1}
Christoph Feichtenhofer, Axel Pinz, and Richard~P. Wildes.
\newblock Spatiotemporal residual networks for video action recognition.
\newblock In {\em NeurIPS}, 2016.

\bibitem{early_multimodal_work2}
Karen Simonyan and Andrew Zisserman.
\newblock Two-stream convolutional networks for action recognition in videos.
\newblock In {\em NeurIPS}, 2014.

\bibitem{early_multimodal_work3}
Guiyu Liu, Jiuchao Qian, Fei Wen, Xiaoguang Zhu, Rendong Ying, and Peilin Liu.
\newblock Action recognition based on 3d skeleton and rgb frame fusion.
\newblock In {\em 2019 IEEE/RSJ International Conference on Intelligent Robots and Systems (IROS)}, pages 258--264, 2019.

\bibitem{slowfast}
Christoph Feichtenhofer, Haoqi Fan, Jitendra Malik, and Kaiming He.
\newblock Slowfast networks for video recognition.
\newblock In {\em CVPR}, 2019.

\bibitem{long_work}
Xiang Long, Chuang Gan, Gerard de~Melo, Xiao Liu, Yandong Li, Fu~Li, and Shilei Wen.
\newblock Multimodal keyless attention fusion for video classification.
\newblock In {\em AAAI}, AAAI'18/IAAI'18/EAAI'18. AAAI Press, 2018.

\bibitem{late_fusion}
Songyang Zhang, Yang Yang, Jun Xiao, Xiaoming Liu, Yi~Yang, Di~Xie, and Yueting Zhuang.
\newblock Fusing geometric features for skeleton-based action recognition using multilayer lstm networks.
\newblock {\em IEEE Transactions on Multimedia}, 20(9):2330--2343, 2018.

\bibitem{residule_conncet}
Christoph Feichtenhofer, Axel Pinz, and Richard~P. Wildes.
\newblock Spatiotemporal residual networks for video action recognition.
\newblock In {\em NeurIPS}, 2016.

\bibitem{ecr}
Pascal Vincent, Hugo Larochelle, Y.~Bengio, and Pierre-Antoine Manzagol.
\newblock Extracting and composing robust features with denoising autoencoders.
\newblock pages 1096--1103, 01 2008.

\bibitem{context}
Deepak Pathak, Philipp Krahenbuhl, Jeff Donahue, Trevor Darrell, and Alexei~A. Efros.
\newblock Context encoders: Feature learning by inpainting.
\newblock In {\em Proceedings of the IEEE/CVF Conference on Computer Vision and Pattern Recognition (CVPR)}, 2016.

\bibitem{beit}
Hangbo Bao, Li~Dong, Songhao Piao, and Furu Wei.
\newblock Beit: Bert pre-training of image transformers.
\newblock In {\em ICLR}, 2022.

\bibitem{bevt}
Rui Wang, Dongdong Chen, Zuxuan Wu, Yinpeng Chen, Xiyang Dai, Mengchen Liu, Yu-Gang Jiang, Luowei Zhou, and Lu~Yuan.
\newblock Bevt: Bert pretraining of video transformers.
\newblock In {\em Proceedings of the IEEE/CVF Conference on Computer Vision and Pattern Recognition (CVPR)}, 12 2021.

\bibitem{vimpac}
Hao Tan, Jie Lei, Thomas Wolf, and Mohit Bansal.
\newblock Vimpac: Video pre-training via masked token prediction and contrastive learning.
\newblock {\em ArXiv}, abs/2106.11250, 2021.

\bibitem{language_few}
Tom~B. Brown, Benjamin Mann, Nick Ryder, Melanie Subbiah, Jared Kaplan, Prafulla Dhariwal, Arvind Neelakantan, Pranav Shyam, Girish Sastry, Amanda Askell, Sandhini Agarwal, Ariel Herbert-Voss, Gretchen Krueger, Tom Henighan, Rewon Child, Aditya Ramesh, Daniel~M. Ziegler, Jeffrey Wu, Clemens Winter, Christopher Hesse, Mark Chen, Eric Sigler, Mateusz Litwin, Scott Gray, Benjamin Chess, Jack Clark, Christopher Berner, Sam McCandlish, Alec Radford, Ilya Sutskever, and Dario Amodei.
\newblock Language models are few-shot learners.
\newblock In {\em NeurIPS}, 2020.

\bibitem{bert}
Jacob Devlin, Ming-Wei Chang, Kenton Lee, and Kristina Toutanova.
\newblock Bert: Pre-training of deep bidirectional transformers for language understanding.
\newblock In {\em North American Chapter of the Association for Computational Linguistics}, 2019.

\bibitem{mae}
Kaiming He, Xinlei Chen, Saining Xie, Yanghao Li, Piotr Dollár, and Ross Girshick.
\newblock Masked autoencoders are scalable vision learners.
\newblock In {\em Proceedings of the IEEE/CVF Conference on Computer Vision and Pattern Recognition (CVPR)}, 2021.

\bibitem{videomae_v2}
Limin Wang, Bingkun Huang, Zhiyu Zhao, Zhan Tong, Yinan He, Yi~Wang, Yali Wang, and Yu~Qiao.
\newblock Videomae v2: Scaling video masked autoencoders with dual masking.
\newblock In {\em Proceedings of the IEEE/CVF Conference on Computer Vision and Pattern Recognition (CVPR)}, pages 14549--14560, June 2023.

\bibitem{MAML}
Chelsea Finn, Pieter Abbeel, and Sergey Levine.
\newblock Model-agnostic meta-learning for fast adaptation of deep networks.
\newblock In {\em Proceedings of the 34th International Conference on Machine Learning}, pages 1126--1135, 2017.

\bibitem{Meta-Networks}
Tsendsuren Munkhdalai and Hong Yu.
\newblock Meta networks.
\newblock {\em Proceedings of machine learning research}, 70, 03 2017.

\bibitem{memory-modelbased}
Adam Santoro, Sergey Bartunov, Matthew~M. Botvinick, Daan Wierstra, and Timothy~P. Lillicrap.
\newblock One-shot learning with memory-augmented neural networks.
\newblock In {\em Proceedings of the International Conference on Machine Learning}, 2016.

\bibitem{prototypical}
Jake Snell, Kevin Swersky, and Richard~S. Zemel.
\newblock Prototypical networks for few-shot learning.
\newblock {\em CoRR}, 2017.

\bibitem{ITANet}
Songyang Zhang, Jiale Zhou, and Xuming He.
\newblock Learning implicit temporal alignment for few-shot video classification.
\newblock 2021.

\bibitem{vit}
Alexander Kolesnikov, Alexey Dosovitskiy, Dirk Weissenborn, Georg Heigold, Jakob Uszkoreit, Lucas Beyer, Matthias Minderer, Mostafa Dehghani, Neil Houlsby, Sylvain Gelly, Thomas Unterthiner, and Xiaohua Zhai.
\newblock An image is worth 16x16 words: Transformers for image recognition at scale.
\newblock 2021.

\bibitem{tsn}
Limin Wang, Yuanjun Xiong, Zhe Wang, Yu~Qiao, Dahua Lin, Xiaoou Tang, and Luc {Val Gool}.
\newblock Temporal segment networks: Towards good practices for deep action recognition.
\newblock In {\em ECCV}, 2016.

\bibitem{deepspeed}
Reza~Yazdani Aminabadi, Samyam Rajbhandari, Ammar~Ahmad Awan, Cheng Li, Du~Li, Elton Zheng, Olatunji Ruwase, Shaden Smith, Minjia Zhang, Jeff Rasley, and Yuxiong He.
\newblock Deepspeed- inference: Enabling efficient inference of transformer models at unprecedented scale.
\newblock In {\em SC22: International Conference for High Performance Computing, Networking, Storage and Analysis}, pages 1--15, 2022.

\end{thebibliography}

\end{document}